\documentclass[conference]{IEEEtran}
\IEEEoverridecommandlockouts
\usepackage{cite}
\usepackage{amsmath,amssymb,amsfonts}
\usepackage[hidelinks,urlcolor=blue]{hyperref} 
\usepackage{graphicx}
\usepackage{textcomp}
\usepackage{xcolor}
\usepackage{gensymb}
\usepackage{algorithm}
\usepackage{algorithmic}
\graphicspath{{figures/}}
\def\BibTeX{{\rm B\kern-.05em{\sc i\kern-.025em b}\kern-.08em
    T\kern-.1667em\lower.7ex\hbox{E}\kern-.125emX}}

    \newcommand{\cng}[1]{\textcolor{black}{#1}}

\begin{document}

\title{Dynamic Watermark Generation for Digital Images using Perimeter Gated SPAD Imager PUFs

\thanks{This article is based on work supported by the National Science Foundation under Grant No. 2442346.  \textcolor{red}{“This is the accepted version at the IEEE MWSCAS'25 Conf. and submitted to the IEEE for possible publication. Copyright may be transferred without notice and this version may be inaccessible.”}}
}

\author{\IEEEauthorblockN{Md Sakibur Sajal and Marc Dandin}
{Department of Electrical and Computer Engineering},\\
{Carnegie Mellon University,}
{Pittsburgh, Pennsylvania, 15213, USA}\\
\\
{email: mdandin@andrew.cmu.edu}   }


\maketitle

\begin{abstract}
Digital image watermarks as a security feature can be derived from the imager's physically unclonable functions (PUFs) by utilizing the manufacturing variations, \textit{i.e.}, the dark signal non-uniformity (DSNU). While a few demonstrations focused on the CMOS image sensors (CIS) and active pixel sensors (APS), single photon avalanche diode (SPAD) imagers have never been investigated for this purpose. In this work, we have proposed a novel watermarking technique using perimeter gated SPAD (pgSPAD) imagers. We utilized the DSNU of three $\mathbf{64 \times 64}$ pgSPAD imager chips, fabricated in a $\mathbf{0.35~\mu m}$ standard CMOS process and analyzed the simulated watermarks for standard test images from publicly available database. Our observation shows that both source identification and tamper detection can be achieved using the proposed source-scene-specific dynamic watermarks with a controllable sensitivity-robustness trade-off. 
\end{abstract}


\section{Introduction}

Digital image watermarking is a popular technique for embedding a digital code or image (the watermark) in another digital image (the host) to identify the source/ownership, ensure copyright protection and verify image integrity~\cite{Sanivarapu2022DigitalTechniques,Malanowska2024DigitalDetection}. Based on the generation method, the watermark can be static or dynamic; the latter provides more security at the cost of complexity~\cite{Keyvanpour2011RobustDomain,Bansal2003DynamicImages}. Based on the embedding process, they can be perceptible or imperceptible~\cite{Singh2020SecureSurvey}. Embedding can be done in the spatial domain by altering the pixel values or in a transform domain, \textit{e.g.}, discrete cosine transform (DCT) by altering the frequency components~\cite{Tao2014RobustReview,Kadhim2019ComprehensiveResearch}.

In recent years, physically unclonable functions (PUFs) have been studied as a viable option for watermark generation, with a special focus on source camera identification~\cite{Okura2017P0A,Kim2016BiometricsSensor,Kim2018CamPUF:Noise,Cao2015CMOSAuthentication} and tamper detection~\cite{Zheng2020AIdentification}. In these works, CMOS image sensor (CIS) or active pixel sensor (APS) PUFs were utilized.

PUFs are physical functions arising from the unavoidable and unclonable manufacturing variations of the fabricated devices~\cite{Zerrouki2022APUFs,Mahalat2022ImplementationIoT}. These functions take \textit{challenges} as their arguments and return unique \textit{responses}, thus creating a challenge-response-pair (CRP) space. Although PUFs can be implemented using ring-oscillators~\cite{Maiti2010ARO-PUF,Rahman2014ARO-PUF:Design}, memories~\cite{Maes2009Low-OverheadPUFs,Liu2017AFunction}, and delay-lines~\cite{Lin2010Low-powerFunctions,Gassend2002SiliconFunctions}, image sensor-based PUFs are an attractive option for a camera application which utilizes the non-idealities associated with the pixel array, \textit{i.e.}, dark signal non-uniformity (DSNU)~\cite{Kim2018CamPUF:Noise}, reset voltage variation~\cite{Zheng2018SecuringAuthentication}, fixed pattern noise (FPN)~\cite{Okura2021Area-EfficientSensor} \textit{etc.}

\begin{figure}
    \centering
    \includegraphics[width=0.9\linewidth]{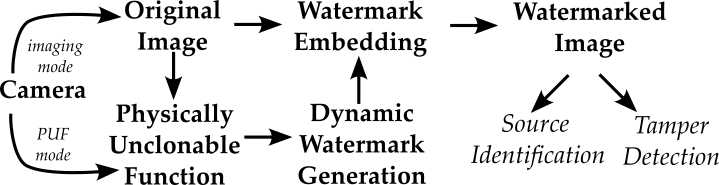}
    \vspace{-8pt}
    \caption{Dynamic watermark generation process from the physically unclonable functions (PUFs) of the image sensor to enable source identification and tamper detection of the captured images. The proposed pgSPAD sensor offers reconfigurable PUFs by utilizing its dark signal non-uniformity (DSNU).}
    \label{fig:overview}
\end{figure}

Figure.~\ref{fig:overview} shows a generic flow diagram to generate dynamically watermarked images for source identification and tamper detection using an imager's PUF. Similarly to the proposal from the authors of Ref.~\cite{Okura2017P0A}, the camera can operate in the imaging mode to capture a scene and then in the PUF mode to generate a signature image \textit{i.e.}, an FPN unique to the camera. By embedding this signature, a static watermarking can be performed. However, by incorporating an image-specific challenge and generating a watermark based on the response, a dynamic watermark can be obtained which can help in tamper detection as well as source identification~\cite{Zheng2020AIdentification}. This watermark can be embedded using any standard steganography technique depending on the application requirement~\cite{Kadhim2019ComprehensiveResearch}. 

In this work, we explored a novel watermarking scheme using the PUFs of perimeter gated single photon avalanche diode (pgSPAD) image sensor arrays~\cite{Sajal2023spadpuf,Sajal2023Challenge-ResponseFunctions}. pgSPADs are a variant of SPADs~\cite{Dandin2025Perimeter-GatedDiodes,Dandin2025OptoelectronicDiodes}, an emerging technology for CMOS image sensor~\cite{Ingle2019HighSensors,Ingle2021PassiveImaging,Ulku2019AFLIM,Wang2024AImaging}, which uses a polysilicon gate over the perimeter junction of the diode to actively modulate the DSNU, a component of the FPN~\cite{Sajal2022Perimeter-GatedProbability}. We have characterized the room temperature DSNU from three prototype chips and generated dynamic watermarks for publicly available gray-scale images. We have demonstrated both source identification and tamper detection using the proposed method. Our watermark generation protocol is designed to compensate for the dark signal variation with temperature. In addition, the proposed method supports a provision for PUF reconfigurability in order to enhance the CRP space.

\begin{figure*}[t]
    \centering
    \includegraphics[width=\linewidth]{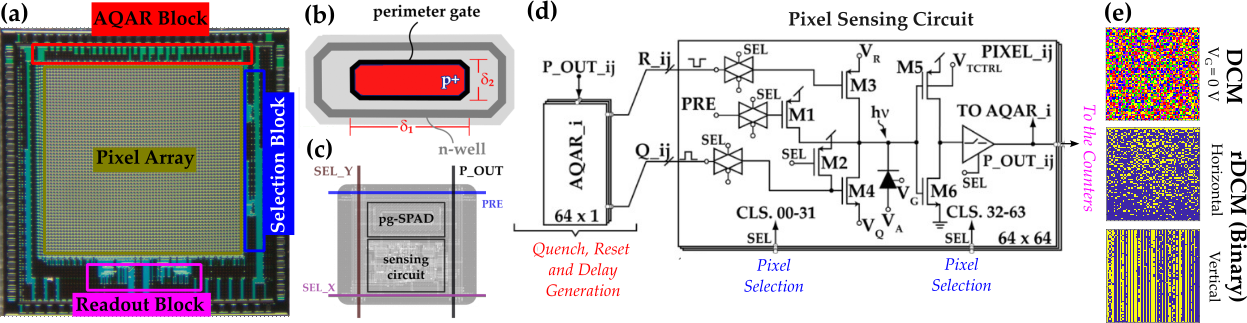}
    \vspace{-10pt}
    \caption{A photomicrograph of a $64\times64$ pgSPAD array surrounded by the peripheral functional blocks, \textit{i.e.}, the pixel selection block, active quench and active reset (AQAR) block and the readout block (a). This $5~mm \times5~mm$ chip was fabricated in a $0.35~\mu m$ standard CMOS process. Layout of the p+/n-well diode showing the perimeter gate and the active area dimensions ($\delta_1=15~\mu m$, $\delta_2=10~\mu m$) (b) and the pixel architecture (c), respectively. Detailed schematic of the pixel sensing circuit (d) and the dark count map (DCM) and relative DCMs (e). An in-depth description of the circuit operation is presented in Ref.~\cite{Sajal2022Perimeter-GatedProbability}.}
    \label{fig:arch}
\end{figure*}

\newpage

\section{Imager Architecture and Operation}

Figure~\ref{fig:arch} (a) shows the photomicrograph of a die with the peripheral functional blocks highlighted. 4,096 pixels ($50~\mu m \times 50~\mu m$), each containing a p+/n-well pgSPAD (see Fig.~\ref{fig:arch} (b)), are tiled in a $64 \times 64$ array with the signal routings shown in Fig.~\ref{fig:arch} (c). The in-pixel sensing circuit is shown in Fig.~\ref{fig:arch} (d). A detailed circuit operation is presented in Ref.~\cite{Sajal2022Perimeter-GatedProbability}. Nevertheless, we briefly describe the operation here.

In reference to Fig.~\ref{fig:arch} (d), the imager is biased at the Geiger mode by applying an anode voltage, $V_A$ ensuring $V_{ex} = V_{R} - V_A > V_{brk}$ where $V_{ex}$ is the excess bias voltage, $V_R\sim V_{DD}$ is the reset voltage, and $V_{brk}$ is the diode's nominal breakdown voltage. When a photon from the scene element impinges on a pixel, an avalanche is probabilistically triggered and a digital pulse is generated at $\mathbf{P_{OUT}}$. The diode is actively quenched to $V_Q$ and reset to $V_R$ by the control signals supplied from the AQAR block. For a controlled exposure time, the output of the selected pixel is accumulated by the counters to be read off-chip. The triggering rate is proportional to the scene intensity (unless dead-time limited non-linearity sets in) and forms an intensity image when the entire array is read~\cite{Dandin2025Perimeter-GatedImagers}.

Similar to photo-generated frames, we can also read dark frames, henceforth referred to as the dark count maps (DCMs), when the imager is operated in the dark. These sensor-specific DCMs were demonstrated as the basis for an imager-based PUF with concealability~\cite{Sajal2023ConcealableImager} and reconfigurability~\cite{Sajal2023Challenge-ResponseFunctions} using different perimeter gate voltages, $V_G$.
In this work, we used the native DCMs \textit{i.e.}, $V_G = 0~V$ to generate binary relative DCMs (rDCMs) by comparing the pixel values horizontally and vertically as shown in Fig.~\ref{fig:arch} (e). Specifically, we assign a pixel to be $1$ in the rDCM if it has a higher dark noise than its corresponding neighbor or a $0$, otherwise.

Demonstrated in Ref.~\cite{Sajal2023ConcealableImager}, even though the DCM pixel values increase exponentially with temperature, these rDCMs tend to remain the same, since a pixel with higher dark noise at room temperature still remains higher than its neighbors at other temperatures. We utilize these native rDCMs as the PUF for the proposed dynamic watermark generation technique.

\section{Proposed Watermark Generation Process}

\begin{figure}
    \centering
    \includegraphics[width=\linewidth]{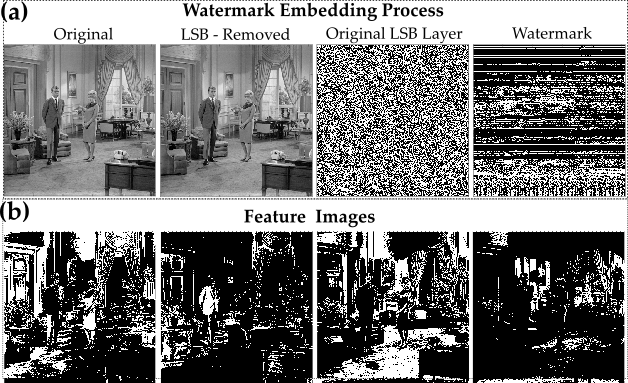}
    \vspace{-10pt}
    \caption{(a) Binary watermark embedding process on the LSB-layer of and (b) example feature images from a 8-bit input gray-scale test image.}
    \label{fig:watermark}
\end{figure}

\begin{figure*}[ht]
    \centering
    \includegraphics[width=\linewidth]{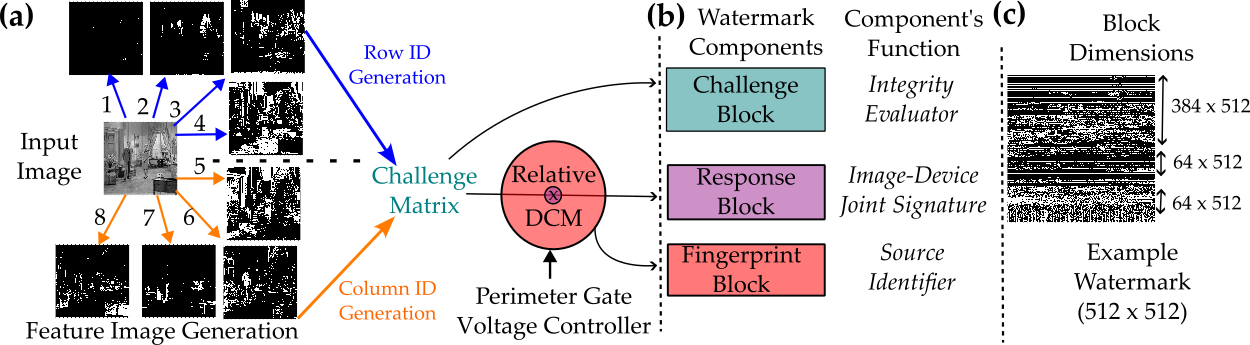}
    \vspace{-10pt}
    \caption{Challenge matrix construction from the binary feature images (a). Perimeter gate voltage can be controlled to alter the PUF function of the imager. Watermark construction by concatenating the challenge, response and device fingerprint blocks (b). An example watermark from a $512\times512$ test image (c). }
    \label{fig:gen}
\end{figure*}

\cng{Figure~\ref{fig:watermark} (a) shows an example watermark embedding process. Since an image looks visually identical to its LSB-removed version, embedding a binary watermark in place of the original LSB layer is one of the popular and simple steganography techniques. It should be noted that more complex and robust methods can be adopted for embedding independent of the proposed watermark generation process~\cite{Kadhim2019ComprehensiveResearch}. The focus of this work is on the watermark generation scheme.} 

\cng{Figure~\ref{fig:watermark} (b) shows examples of binary feature images ($FI$'s) for extracting image information to use in the watermark generation process. Assuming an 8-bit image, $I\in[0,255]$, we generated $FI$'s by quantizing the pixel values as}
\begin{equation}
\label{eq:sth}
    FI_i = sign(sign(\frac{M}{L}\times i -I)+1)-sign(\sum_{k=0}^{i-1} FI_{k})
\end{equation}
Here, $M = 256$ is the range of the pixel values and $L = 8$ is the total number of $FI$'s with $FI_0$ being an all-zeros matrix. Eq.~\ref{eq:sth} is a single thresholding method for assigning a bit ($1$ or $0$) to a pixel if it is above or below a certain value, respectively. 

\textbf{Challenge Formation}: Figure~\ref{fig:gen} (a) shows the combining process of the $FI$'s into a challenge matrix, \cng{\textit{i.e.}, a set of pixel addresses to inquire for their rDCRs}. By concatenating the bits from the $FI$'s at each pixel location, binary numbers are created which represent the challenge addresses for the PUF. We used the first four $FI$'s for the row address and the last four $FI$'s for the column address generation. These challenge addresses become a part (challenge block) of the final watermark (see Fig.~\ref{fig:gen} (b)) to function as a quick integrity evaluator. Any alteration of the pixel values greater than the sensitivity threshold, $M/L$ will affect two $FI$'s \cng{simultaneously} and consequently the challenge block of the watermark.

\textbf{Response Formation}: The challenge matrix, $C$ is passed to the PUF as an argument for an image-device-specific response generation as
\begin{equation}
    R = PUF_{V_G}(C).
\end{equation}
Here, $PUF_{V_G}$ is the PUF associated with the applied gate voltage. For this work, we used the native PUF, \textit{i.e.}, $V_G=0~V$. 

In essence, $PUF_{V_G}$ reads the rDCMs from the pixel addresses associated with $C$ and creates the response block (see Fig.~\ref{fig:gen} (b)) for the watermark. This joint signature block functions as the source-scene-pair verifier. Any discrepancies in the image and/or the source camera information will be reflected in this block. For instance, if a doctored image claims to be \textit{originally} generated from a particular camera, a verifier can use the rDCMs \cng{of the alleged camera and the $FI$'s of the doctored image} to verify the claim. In PUF based applications, an enrolled device's PUF information (rDCM in this case) is usually stored in a secured database with a trusted third party for response verification as needed.

\begin{figure}
    \centering
    \includegraphics[width=\linewidth]{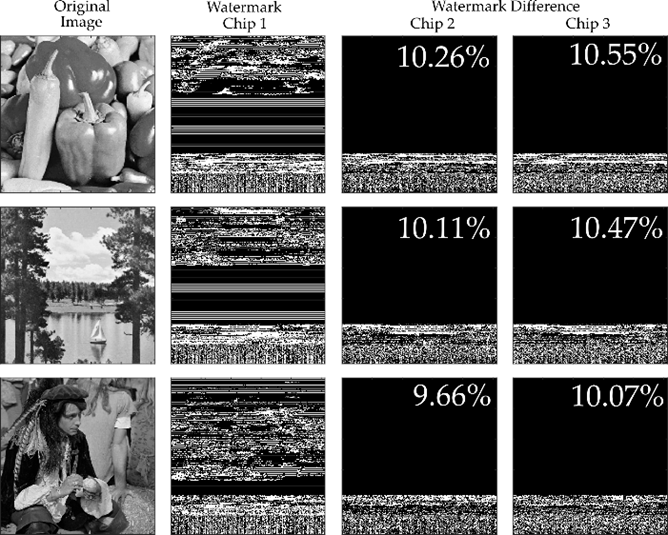}
    \vspace{-10pt}
    \caption{Proposed watermarking process verification using three pgSPAD imager chips and three standard test images from public database.}
    \label{fig:source}
\end{figure}

\textbf{Fingerprint Formation}:
We performed a bit-wise exclusive-OR operation on the horizontal and the vertical rDCMs to create a fingerprint block (see Fig.~\ref{fig:gen} (b)) for the watermark. This fingerprint carries the device identity and the XOR operation masks the actual rDCM values to maintain their secrecy. This block functions as a camera identifier from a registered database in the event of metadata unavailability. Fingerprint block is the sole attribute of the source camera.

Finally, Fig~\ref{fig:gen} (c) shows an example watermark generated following the proposed method from a $512\times512$ gray-scale test image. The dimensions of the three component blocks are also provided. The dimensions of the blocks can be easily adapted for any another format or image size.

\section{Experimental Results and Discussions}
\subsection{Source Identification}

Figure~\ref{fig:source} shows three watermarks generated for three different images from chip 1. The challenge blocks and the response blocks are visibly different from each other since these blocks are specifically image dependent. However, the fingerprint blocks are identical as they are from the same source, \textit{i.e.}, chip 1. This verifies the image-specific dynamic nature of the watermarks. 

We also show the watermark difference for the same images but generated from other chips. By design, the response and the fingerprint blocks will perform the differentiation now since the challenge block is identical for the same image. We see an average of $10\%$ mismatch coming from the response and the fingerprint blocks. This verifies the source-specific dynamic nature of the watermarks.

\subsection{Tamper Detection}
\begin{figure}
    \centering
    \includegraphics[width=\linewidth]{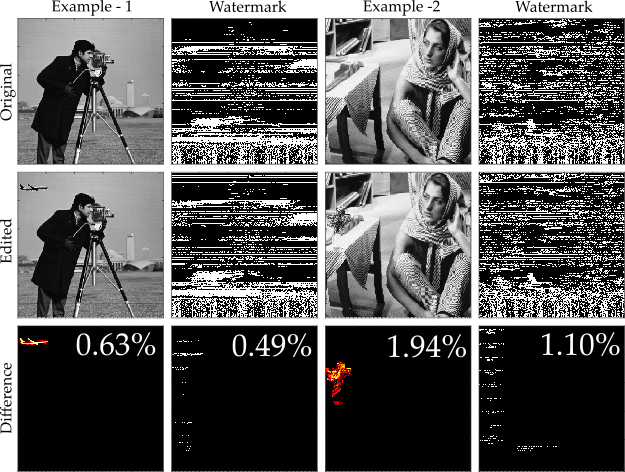}
    \vspace{-10pt}
    \caption{Tamper detection from the proposed watermark generation technique. Two original test images were edited and watermarks were generated for each to compare. The proposed technique produces nonlocal variations in response to local alterations to the original images as indicated in the bottom row.}
    \label{fig:tamper}
\end{figure}
Fig.~\ref{fig:tamper} shows two sets of original and edited images (the differences are highlighted) \textit{claiming} to be captured by chip 1. Despite the edited images being able to copy the fingerprint block, the watermark difference in the challenge and the response blocks still gives them away. We see that a $0.63\%$ change in the first image resulted in a $0.49\%$ change in the watermark, while a $1.94\%$ change in the second image resulted in a $1.10\%$ change in the corresponding watermark. It should be noted that a localized change in the pixel values results in a nonlocal spread-out change in the watermark; a built-in feature of the proposed watermark generation technique to deter an adversary from predicting the affected pixel locations on the new watermark. As long as the adversary changes a pixel value by the $FI$ threshold, there will be a maximum of $L$ locations affected by that. 

We can define the sensitivity, $S$ of the system as
\begin{equation}
    S = \frac{\%~change ~of ~the~watermark}{\%~change~of~the~input~image}.
\end{equation}
By increasing the number of feature images, the sensitivity of the system can be increased at the cost of computation.

\subsection{Robustness Analysis}
\begin{figure}
    \centering
    \includegraphics[width=\linewidth]{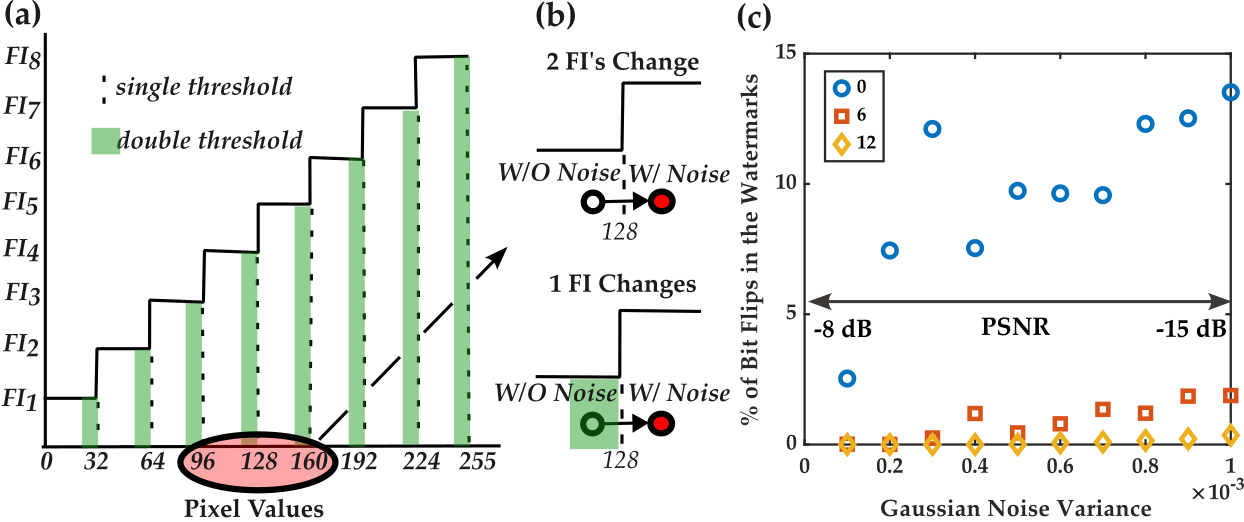}
    \vspace{-10pt}
    \caption{Robustness analysis of the proposed technique. By implementing a double threshold for the $FI$ change, we can trade sensitivity for robustness.}
    \label{fig:robust}
\end{figure}
The $FI$ thresholding trades sensitivity with robustness. Unlike complex spectral feature-based methods~\cite{Zheng2020AIdentification}, simple intensity-based $FI$'s are more susceptible to noise. To address that, we can employ a double threshold (DTH)-based quantization approach for $FI$ generation by creating a desensitized range around the single-threshold boundary (see the green shaded regions in Fig.~\ref{fig:robust} (a)). Pixel values corresponding to the desensitized regions contribute to both adjacent $FI$'s. As a result, changes in pixel values due to noise do not alter adjacent $FI$'s simultaneously. Instead, one $FI$ changes when the noise magnitude is comparable to the threshold overlap as shown in Fig.~\ref{fig:robust} (b). 

By changing the overlap width, noise immunity can be increased at the expense of sensitivity (see Fig.~\ref{fig:robust} (c)). We have added different amount of Gaussian noise to a test image with a peak signal-to-noise ratio (PSNR) ranging from $-8~dB$ to $-18~dB$. For each case, the double thresholding was applied for $FI$ generation with $0$, $6$ and $12$ overlaps. We see that the percent bit-flips with respect to the reference watermark (generated from the noise-free input image) decreases with the increase of the double thresholding overlap width. This idea is analogous to the application of Schmitt triggers as comparators for better noise immunity.

\begin{table}
    \centering
        \caption{Qualitative Comparison with Other PUF-based Methods}
    \label{tab:comp} 
    \begin{tabular}{|c|c|c|c|c|} \hline
         &  2018\cite{Kim2018CamPUF:Noise}&  2018\cite{Zheng2018SecuringAuthentication}&  2021\cite{Okura2021Area-EfficientSensor} & This Work\\ \hline
         PUF Source&  DSNU&  Reset V.&  FPN & DSNU\\
          Source Id. &  Yes&  Yes&  Yes &Yes\\
         Tamp. Det. &  No&  Yes&  No & Yes\\
         Feat. Extract&  -&  DCT&  - & DTH\\
\hline
    \end{tabular}

\end{table}

\section{Conclusions}
In this work, we have presented a novel watermark generation method leveraging the physically unclonable functions (PUFs) of perimeter gated single photon avalanche diode (pgSPAD) imagers. Being an emerging imaging technology, SPAD cameras have the potential to capture and watermark digital images to ensure tamper resistance and trusted provenance similar to the CIS or APS PUF-based watermarking protocols shown in Table~\ref{tab:comp} \cng{with comparable circuit overhead and PUF database infrastructure.} We have demonstrated a temperature resilient method of utilizing the DSNU of the sensor for extracting the PUFs and generate dynamic watermarks. The focus of the current work was to explore an embedding process agnostic watermarking scheme for native SPAD imagers. A more sophisticated method can be adopted by incorporating perimeter gate voltage as a part of the challenge response pair generation~\cite{Sajal2023Challenge-ResponseFunctions}, which is left as the future work.

\newpage

\bibliographystyle{IEEEtranDOI}
\bibliography{main}

\end{document}